\documentclass{vgtc}                          % final (conference style)
\graphicspath{{figures/}{pictures/}{images/}{./}} % where to search for the images

\usepackage{times}                     % we use Times as the main font
\usepackage{tabu}                      % only used for the table example
\usepackage{booktabs}                  % only used for the table example
\usepackage{lipsum}                    % used to generate placeholder text
\usepackage{mwe}                       % used to generate placeholder figures

\usepackage{mathptmx}                  % use matching math font
\usepackage{multirow}
\usepackage{multicol}

\onlineid{1043}

\vgtccategory{Poster}

\vgtcinsertpkg

\title{Creating Virtual Environments with 3D Gaussian Splatting: \\A Comparative Study}

\author{Shi Qiu\thanks{e-mail: shiqiu@cse.cuhk.edu.hk\\{This work was supported by The Chinese University of Hong Kong (Project No.: 4055212); and in part by the Research Grants Council of the Hong Kong Special Administrative Region, China (Project No.: T45-401/22-N).}}, \quad Binzhu Xie, \quad Qixuan Liu, \quad Pheng-Ann Heng\\ %
        \scriptsize Department of Computer Science and Engineering, The Chinese University of Hong Kong\\
        \scriptsize Institute of Medical Intelligence and XR, The Chinese University of Hong Kong}

\teaser{
  \centering
  \includegraphics[alt={An overview of 3DGS-based approaches to VE creation.}, width=0.82\linewidth]{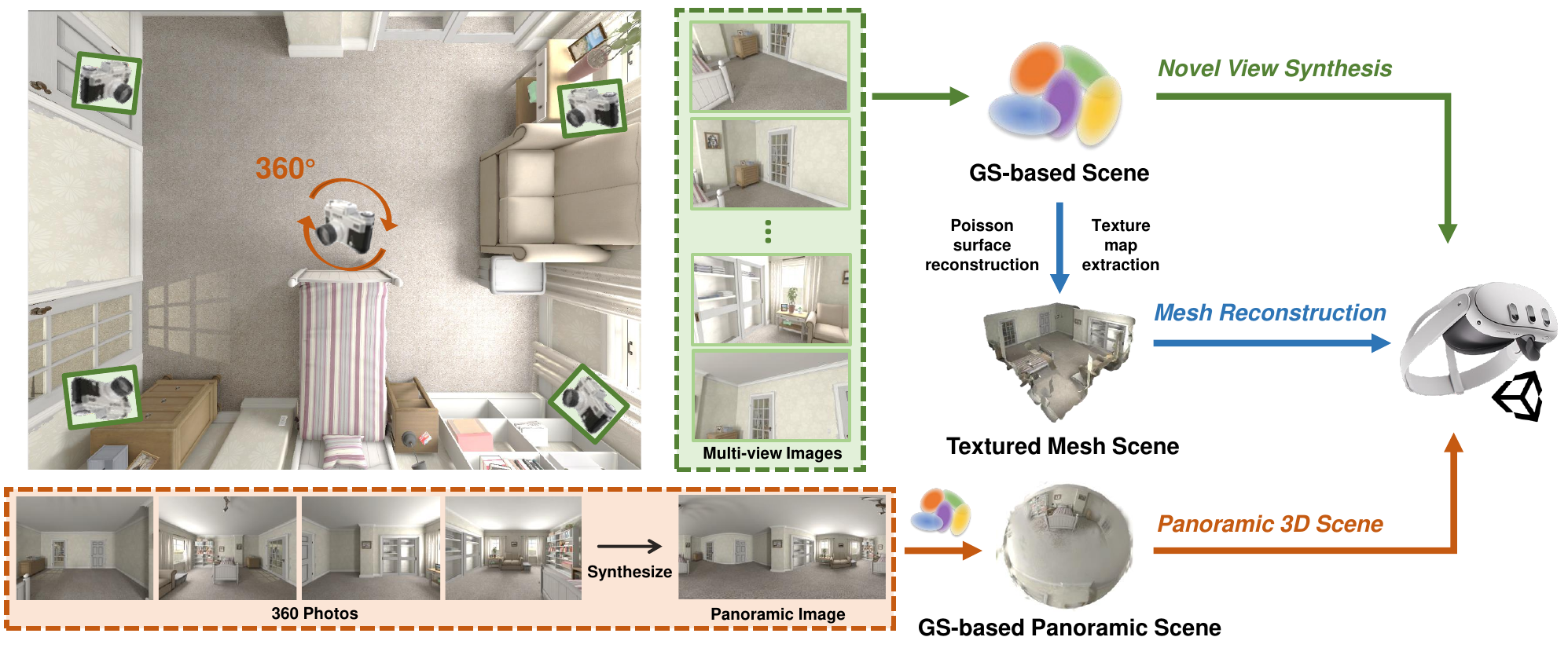}
  
  \caption{We develop three different 3D Gaussian Splatting approaches for virtual environment creation, where the corresponding pipelines are illustrated in green, blue, and orange colors, respectively. More implementation details are introduced in Sec.~\ref{sec:impl}.}
  \label{fig:teaser}
}

\abstract{3D Gaussian Splatting (3DGS) has recently emerged as an innovative and efficient 3D representation technique. While its potential for extended reality (XR) applications is frequently highlighted, its practical effectiveness remains underexplored. In this work, we examine three distinct 3DGS-based approaches for virtual environment (VE) creation, leveraging their unique strengths for efficient and visually compelling scene representation. By conducting a comparable study, we evaluate the feasibility of 3DGS in creating immersive VEs, identify its limitations in XR applications, and discuss future research and development opportunities.% filler text. Replace with your abstract.
} % end of abstract


\begin{document}

%% The ``\maketitle'' command must be the first command after the
%% ``\begin{document}'' command. It prepares and prints the title block.

%% the only exception to this rule is the \firstsection command
%\firstsection{Introduction}
\maketitle

\section{Introduction} %for journal use above \firstsection{..} instead

Virtual environments (VEs) enable users to interact seamlessly with digital worlds across a wide range of extended reality (XR) applications. To deliver truly immersive experiences, XR applications require VEs to be highly detailed, visually realistic, and responsive. However, creating such environments remains a significant challenge, particularly in achieving a balance between rendering efficiency, scalability, and visual fidelity. Given the performance constraints of XR devices like head-mounted displays (HMDs), the creation and rendering of VE become crucial for user immersion.

Traditional VE creation techniques, such as 3D modeling and photogrammetry, rely heavily on polygonal meshes and manually crafted assets. While effective, these methods are time-consuming, labor-intensive, and struggle to represent complex scenes. Recent advances in neural rendering have introduced to synthesize novel views from sparse input data. For instance, Neural Radiance Fields (NeRF)~\cite{mildenhall2021nerf} utilize multi-layer perceptrons (MLPs) for implicit scene encoding but face computational bottlenecks in real-time applications. More recently, 3D Gaussian Splatting (3DGS)~\cite{kerbl20233d} represents scenes with explicit 3D Gaussian points, offering practical advantages: faster and more scalable rendering, high visual fidelity, and straightforward scene manipulation~\cite{qiu2025advancing}. These features make the 3DGS techniques promising for creating immersive VEs.

In this work, we explore the use of advanced 3DGS techniques in VE creation for XR applications, particularly in balancing high visual quality with the performance constraints of real-time rendering on HMDs. As illustrated in Fig.~\ref{fig:teaser}, we introduce and evaluate three distinct 3DGS-based approaches for VE creation: (i) the novel view synthesis approach, (ii) the mesh reconstruction approach, and (iii) the panoramic 3D scene approach. By conducting a comparative study, we aim to evaluate the feasibility and effectiveness of 3DGS in creating immersive VEs, identify its limitations in XR applications, and provide insights for future research and development.

\section{Implementation}
\label{sec:impl}

For the \textbf{novel view synthesis approach}, we start by capturing multiview images of the target environment and then proceed to estimate camera parameters and a sparse point cloud using COLMAP~\cite{schoenberger2016mvs}. Subsequently, we leverage the original 3DGS framework~\cite{kerbl20233d} to train 3D Gaussian representations of the scene. Once trained, the 3D Gaussian data is imported into Unity to synthesize the novel view of the GS-based scene by decoding the learned information. We use an open-source shader implementation\footnote{https://github.com/clarte53/GaussianSplattingVRViewerUnity} for the real-time rendering of GS-based scenes in XR.

\begin{figure}
    \centering
    \includegraphics[alt={VE results generated from different 3DGS approaches.}, width=.82\linewidth]{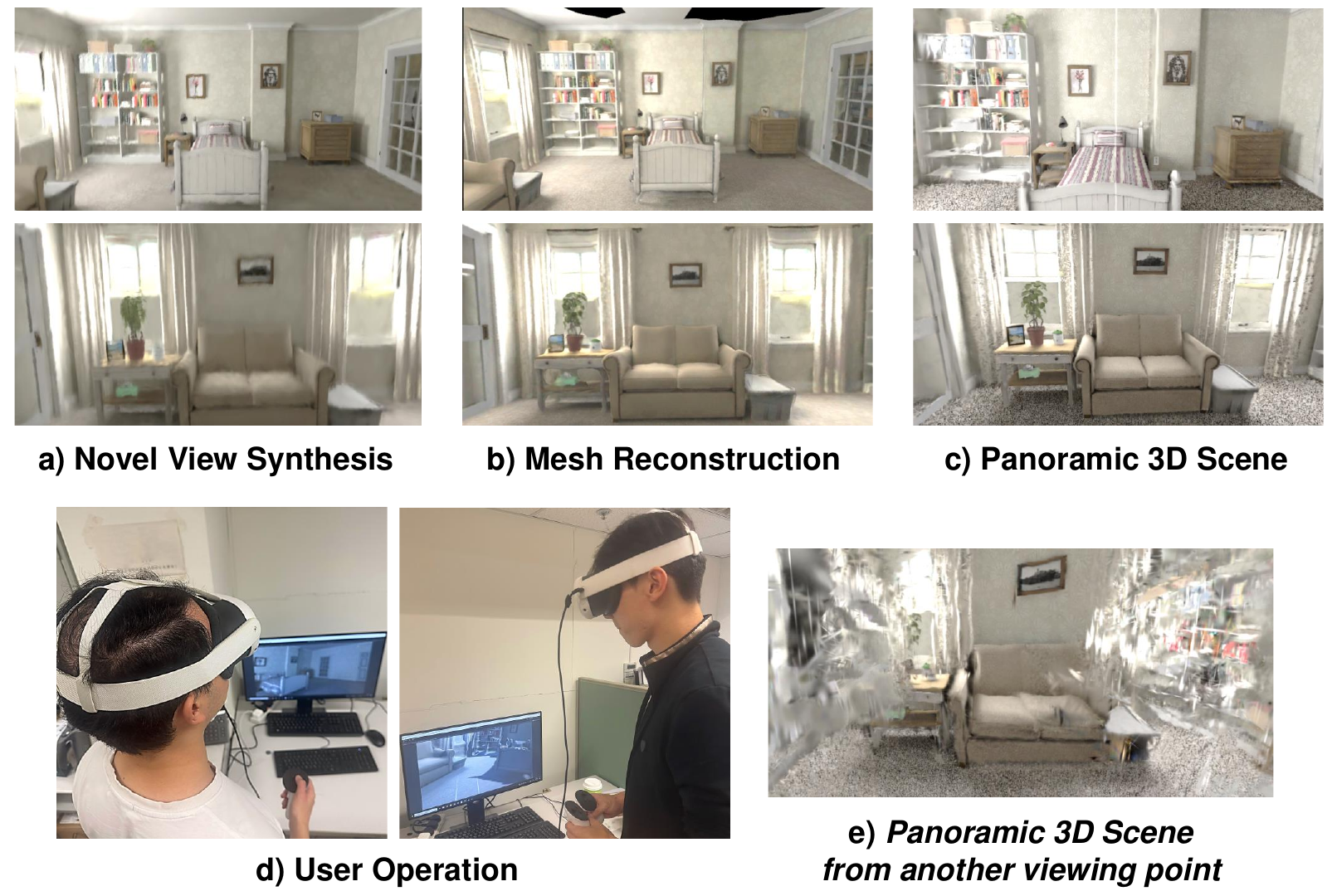}
    \caption{Qualitative comparisons of VE rendering results.}
    \label{fig:qcom}
\end{figure}

In the \textbf{mesh reconstruction approach}, our primary focus lies in preserving fine details within the reconstructed scene. Building upon recent SuGaR~\cite{sugar} framework, we apply regularization constraints to train more evenly distributed Gaussians across the scene’s surface. Given enhanced 3DGS representations, we sample the scene to extract information about visible points. Subsequently, the Poisson surface reconstruction algorithm~\cite{kazhdan2006poisson} is utilized to generate a triangular mesh based on the positions and normals of the sampled points. We further texture the mesh-based scene given the cues from trained Gaussians~\cite{frost}, importing into Unity for XR use. 

In the \textbf{panoramic 3D scene approach}, our initial step is to generate a panoramic image of the target scene by synthesizing the multiview images captured from a fixed location. Like the DreamScene360~\cite{zhou2025dreamscene360} pipeline, we use a pre-trained monocular depth estimator alongside an optimizable geometric field to establish a consistent scene geometry structure. To address unobserved areas, pseudo-views and deformed Gaussian points are generated, leading to a high-quality panoramic scene represented in 3DGS. Finally, this GS-based panoramic scene is imported and rendered in Unity via a similar process as used in the novel view synthesis approach.

% \binzhu{Reviewers: ``how does it perform to Panorama + Monodepth360? if the user can only look around and is not able to change their standing point?''
% I thick monodepth360 is just a depth estimator, cannot be compared}

\section{Evaluation}
\label{sec:evl}

We render the created VE models on an NVIDIA RTX 4090 GPU, with a Meta Quest 3 as the HMD for testing. As shown in Fig.~\ref{fig:qcom} (a-c), we qualitatively compare the rendered first-person views. All three approaches deliver comparable visual fidelity, with some regular artifacts such as motion blur observed in (a) and (c), and reconstruction incompleteness noted in (b). In terms of rendering speed, all approaches achieve approximately 70 FPS, demonstrating the feasibility of 3DGS for real-time XR applications.

%\binzhu{User Details:   How old are the people? how used have they been to VR? do they know Gaussian Splatting.}
We conduct a user study with 8 participants to evaluate the sense of presence and immersion within the created VEs, using the \emph{Slater, Usoh, and Steed} (SUS) questionnaire~\cite{usoh2000using}. As shown in Tab.~\ref{tab:user}, the novel view synthesis approach achieves the highest SUS Count and Mean metrics. It also scores the highest in five out of six SUS questions. Users report that this approach maintained consistent visual quality as they moved through the VE, enhancing spatial awareness. By contrast, the other two approaches only present high-fidelity details in specific views, suffering from spatial distortions and artifacts that negatively impact user immersion when moving around.

\begin{table}
  \caption{Summary of user study results ($n=8$). The study comprises 7 male and 1 female participants with an average age of 25. All participants have limited experience with VR or 3DGS. We use the SUS questionnaire \cite{usoh2000using} to quantitatively compare users' presence scores (mean $\pm$ std; higher values indicate a greater sense of presence).}
  \label{tab:user}
  \centering%
  \renewcommand{\arraystretch}{1.1} %adjust line spacing
  \resizebox{0.82\linewidth}{!}{
  \begin{tabu}{%
    c|ccc
    }
  \toprule
  \multirow{2}{*}{Score} & \multicolumn{3}{c}{\emph{Virtual environments created using the approaches of}} \\
   &\textbf{\emph{Novel View Synthesis}}
  &\textbf{\emph{Mesh Reconstruction}} 
  &\textbf{\emph{Panoramic 3D Scene}}\\\midrule
    {Q1} &\textbf{5.5$\pm$1.4}&4.8$\pm$0.8&2.1$\pm$1.0\\
    {Q2} &\textbf{4.3$\pm$1.3}&4.0$\pm$1.4&1.8$\pm$1.0\\
    {Q3} &5.1$\pm$2.0&\textbf{5.3$\pm$1.8}&2.8$\pm$1.7\\
    {Q4} &\textbf{5.6$\pm$1.6}&4.8$\pm$1.6&2.0$\pm$1.3\\
    {Q5} &\textbf{4.8$\pm$1.8}&4.3$\pm$1.3&1.9$\pm$0.8\\
    {Q6} &\textbf{4.0$\pm$2.0}&3.8$\pm$1.8&1.1$\pm$0.4\\\midrule
    \textbf{SUS Count} &\textbf{2.8$\pm$1.8}&1.8$\pm$2.4&0.0$\pm$0.0\\
    \textbf{SUS Mean} &\textbf{4.9$\pm$1.2}&4.5$\pm$1.1&1.9$\pm$0.9\\
  \bottomrule
  \end{tabu}%
  }
\end{table}

\section{Discussion}
In addition to the evaluations in Sec.~\ref{sec:evl}, we identify several challenges during implementation and experimentation. The \textbf{novel view synthesis approach} stands out due to its higher rendering efficiency, consistency, and editability. However, while the learned 3DGS representation captures geometric details, it offers only basic collision and occlusion modeling in XR environments when treating all Gaussians uniformly. To enhance object-level immersion and interaction, a more robust mechanism is needed to identify and prioritize Gaussian points associated with specific objects. The \textbf{mesh reconstruction approach} is a downstream application of 3DGS, which has received limited research attention. Our additional tests find that 3DGS-based mesh reconstruction could \emph{hardly} compete with traditional TSDF~\cite{izadi2011kinectfusion} methods, particularly when reconstructing distant objects. Despite these shortcomings, mesh-based representations remain foundational for XR development, and further research into 3DGS-driven mesh reconstruction holds significant potential. The \textbf{panoramic 3D scene approach} inherits the strengths of traditional 360-degree panoramas, making it ideal for applications like virtual tours. However, it is less effective for XR applications requiring free movement in larger spaces, as noted in the user study results (Tab.~\ref{tab:user}). Visual artifacts frequently occur when users change their viewing positions (Fig.~\ref{fig:qcom}(c, e)). Even slight changes in the user's standing position can significantly impact the display performance, as the panoramic image is constrained to a fixed viewpoint. Future work may focus on integrating multiview panoramas to create high-fidelity, 3DGS-based 360-degree scenes that support dynamic user movement.

\section{Conclusion}

This work explores three 3DGS-based approaches for virtual environment creation, comparing their practical effectiveness through system evaluations and user studies. Our findings highlight the advantages of 3DGS in delivering efficient and visually compelling VEs. We also identify challenges unique to each approach, such as limited object-level interaction and spatial distortions in certain scenarios. These insights underline the potential of 3DGS for XR applications while pointing to future research directions.

%% if specified like this the section will be committed in review mode
% \acknowledgments{This work was supported by The Chinese University of Hong Kong (Project No.: 4055212); and in part by the Research Grants Council of the Hong Kong Special Administrative Region, China (Project No.: T45-401/22-N).}

\bibliographystyle{abbrv}

\bibliography{template}

\begin{thebibliography}{10}

\bibitem{sugar}
A.~Gu{\'e}don and V.~Lepetit.
\newblock Sugar: Surface-aligned gaussian splatting for efficient 3d mesh reconstruction and high-quality mesh rendering.
\newblock {\em CVPR}, 2024.

\bibitem{frost}
A.~Guédon and V.~Lepetit.
\newblock Gaussian frosting: Editable complex radiance fields with real-time rendering.
\newblock In {\em ECCV}, 2024.

\bibitem{izadi2011kinectfusion}
S.~Izadi et~al.
\newblock Kinectfusion: real-time 3d reconstruction and interaction using a moving depth camera.
\newblock In {\em ACM UIST}, 2011.

\bibitem{kazhdan2006poisson}
M.~Kazhdan, M.~Bolitho, and H.~Hoppe.
\newblock Poisson surface reconstruction.
\newblock In {\em Eurographics/ACM SGP}, volume~7, 2006.

\bibitem{kerbl20233d}
B.~Kerbl, G.~Kopanas, T.~Leimk{\"u}hler, and G.~Drettakis.
\newblock 3d gaussian splatting for real-time radiance field rendering.
\newblock {\em ACM Trans. Graph.}, 42(4):139--1, 2023.

\bibitem{mildenhall2021nerf}
B.~Mildenhall, P.~P. Srinivasan, M.~Tancik, J.~T. Barron, R.~Ramamoorthi, and R.~Ng.
\newblock Nerf: Representing scenes as neural radiance fields for view synthesis.
\newblock {\em Communications of the ACM}, 65(1):99--106, 2021.

\bibitem{qiu2025advancing}
S.~Qiu, B.~Xie, Q.~Liu, and P.-A. Heng.
\newblock Advancing extended reality with 3d gaussian splatting: Innovations and prospects.
\newblock In {\em IEEE AIxVR}, 2025.

\bibitem{schoenberger2016mvs}
J.~L. Sch\"{o}nberger, E.~Zheng, M.~Pollefeys, and J.-M. Frahm.
\newblock Pixelwise view selection for unstructured multi-view stereo.
\newblock In {\em ECCV}, 2016.

\bibitem{usoh2000using}
M.~Usoh, E.~Catena, S.~Arman, and M.~Slater.
\newblock Using presence questionnaires in reality.
\newblock {\em Presence}, 9(5):497--503, 2000.

\bibitem{zhou2025dreamscene360}
S.~Zhou et~al.
\newblock Dreamscene360: Unconstrained text-to-3d scene generation with panoramic gaussian splatting.
\newblock In {\em ECCV}, 2024.

\end{thebibliography}
\end{document}